\documentclass{article}

\usepackage{microtype}
\usepackage{graphicx}
\usepackage{subcaption}
\usepackage{booktabs} 
\usepackage{array} 

\usepackage{hyperref}

\usepackage[accepted]{icml2026}

\usepackage{amsmath}
\usepackage{amssymb}
\usepackage{mathtools}
\usepackage{amsthm}

\usepackage[capitalize,noabbrev]{cleveref}

\theoremstyle{plain}

\theoremstyle{definition}

\theoremstyle{remark}

\usepackage[textsize=tiny]{todonotes}

\icmltitlerunning{MedCase-Structured: A Text-to-FHIR Dataset for Benchmarking Diagnostic Reasoning in Clinically Realistic EHR Settings}

\usepackage{listings}
\usepackage{xcolor}

\lstdefinelanguage{json}{
    basicstyle=\ttfamily\tiny,
    lineskip=-1pt, 
    string=[s]{"}{"},
    stringstyle=\color{blue!60!black},
    comment=[l]{//},
    commentstyle=\color{gray},
    literate=
        {:}{{{\color{red!70!black}{:}}}}{1}
        {,}{{{\color{red!70!black}{,}}}}{1}
        {\{}{{{\color{black}{\{}}}}{1}
        {\}}{{{\color{black}{\}}}}}{1}
        {[}{{{\color{black}{[}}}}{1}
        {]}{{{\color{black}{]}}}}{1},
}

\usepackage{tcolorbox}
\tcbuselibrary{skins, breakable}

\newtcolorbox{prompt}[1][]{
    colback=gray!5,
    colframe=gray!50,
    fonttitle=\bfseries\small,
    title=#1,
    breakable,                
    enhanced,
    boxrule=0.5pt,
    left=6pt,
    right=6pt,
    top=4pt,
    bottom=4pt,
}

\begin{document}

\twocolumn[
  \icmltitle{MedCase-Structured: A Text-to-FHIR Dataset for Benchmarking Diagnostic Reasoning in Clinically Realistic EHR Settings}



  \icmlsetsymbol{equal}{*}

  \begin{icmlauthorlist}
    \icmlauthor{Valentina Bui Muti}{sys}
    \icmlauthor{Eug\'enie Dulout}{sys}
    \icmlauthor{Ziquan Fu}{sys}
  \end{icmlauthorlist}

  \icmlaffiliation{sys}{System Inc., New York, NY, USA}

  \icmlcorrespondingauthor{Valentina Bui Muti}{valentina@system.com}
  \icmlcorrespondingauthor{Ziquan Fu}{frank@system.com}

  \icmlkeywords{large language models, clinical decision support, synthetic data generation, electronic health records, clinical reasoning, medical benchmarking, structured clinical data}

  \vskip 0.3in
]



\printAffiliationsAndNotice{}  

\begin{abstract}
Large language models (LLMs) show promise for clinical reasoning and decision support, but evaluation in structured, electronic health record-congruent settings remains limited. Existing benchmarks often rely on static datasets or unstructured inputs that do not reflect the interoperable data formats used in clinical systems. We introduce a reusable pipeline for generating terminology-grounded HL7 FHIR R4 bundles from unstructured text, enabling controllable evaluation of clinical decision support systems over structured inputs. The pipeline combines staged LLM generation with terminology-grounded validation and repair to eliminate hallucinated codes and enforce structural and semantic consistency. Applying this approach to MedCaseReasoning, we construct \textbf{MedCase-Structured}, a synthetic dataset of 1,732 FHIR bundles derived from clinician-authored diagnostic cases, producing complete, valid bundles for 97.1\% of attempted cases. Evaluation on \textbf{MedCase-Structured} reveals consistently lower diagnostic accuracy for LLMs on structured FHIR inputs than with plain text, highlighting the importance of deployment-aligned benchmarking.
\end{abstract}

\section{Introduction}

\begin{figure*}[t]
\centering
\includegraphics[width=\textwidth]{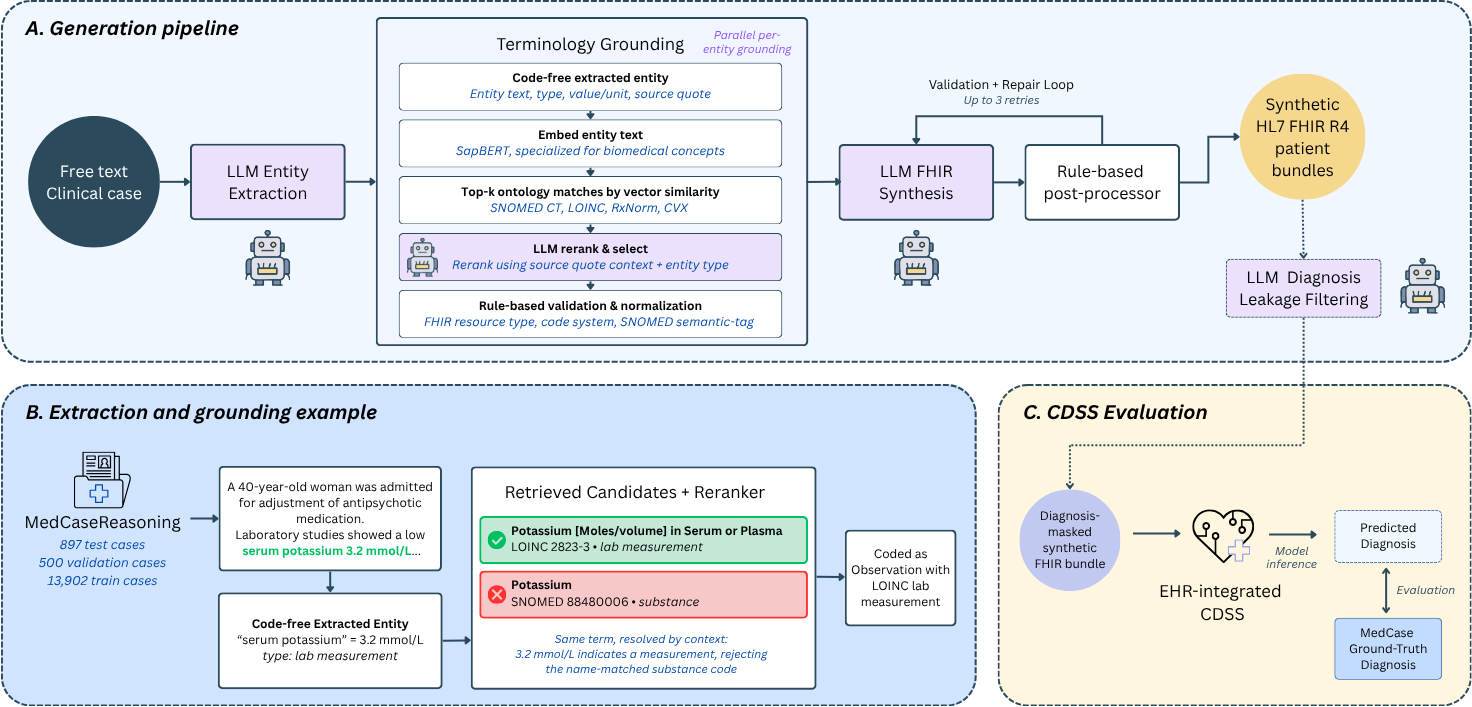}
\caption{Overview of MedCase-Structured. (A) Free-text clinical cases are converted into terminology-grounded HL7 FHIR R4 bundles. (B) An example MedCaseReasoning \citep{wu_medcasereasoning_2025} case shows extraction, ontology-aware grounding, and rejection of a name-matched but semantically invalid code. (C) Diagnosis-masked bundles are used for EHR-congruent CDSS evaluation against ground-truth diagnosis.}
\label{fig:pipeline}
\end{figure*}

Large language models (LLMs) have demonstrated promising capabilities across a range of clinical reasoning and decision support tasks \citep{shool_systematic_2025, mansoor_reasoning_2025}, motivating their use in clinical decision support systems (CDSS). The richness of patient data captured in electronic health records (EHRs) makes them a valuable input source for LLM-based CDSS. However, EHR data are heterogeneous and largely unstructured \citep{li_scoping_2024}, making it challenging to effectively incorporate full patient context into LLM-based pipelines. As LLM-based CDSS become more prevalent, rigorous testing in structured, clinically meaningful, and deployment-aligned settings is essential.

Evaluating EHR-based CDSS tools presents two key challenges. First, real patient data are protected by strict privacy regulations, limiting access and reproducibility \citep{li_scoping_2024}. Second, evaluation inputs must reflect the structure and standards of real clinical systems. Modern healthcare infrastructure increasingly relies on HL7's Fast Healthcare Interoperability Resources (FHIR) \citep{noauthor_overview_nodate} for representing and exchanging patient data. While datasets such as MIMIC-IV \citep{johnson_mimic-iv_2023} are widely used for benchmarking clinical models \citep{li_scoping_2024}, they are restricted to specific care settings and do not natively preserve EHR interoperability structures. Although derived representations such as MIMIC-IV-FHIR \citep{bennett_mimic-iv_2023} map these into FHIR format, they are retrospective transformations rather than outputs of deployed clinical systems. Additionally, it is meaningfully different to evaluate LLM based CDSS on FHIR-based inputs versus textual inputs, as understanding structured and coded information requires higher reasoning ability because of the absence of semantically continuous context that the LLMs are primarily pre-trained on. Recent work shows that both input representation and evaluation protocols significantly influence LLM performance in clinical tasks  \citep{shool_systematic_2025, navarro_evaluation_2026, yang_ehrstruct_2026}, emphasizing the need for standardized, deployment-aligned benchmarks. Similarly, studies of FHIR-based systems highlight the difficulty of reasoning over structured patient data and the lack of realistic evaluation benchmarks \citep{lee_fhir-agentbench_2025}. 

These challenges highlight the need for publicly available synthetic clinical data that can represent patient context in EHR-congruent formats while supporting controlled evaluation. Tools such as Synthea \citep{walonoski_synthea_2018} generate realistic patient records while bypassing privacy concerns and supporting export in FHIR-compatible formats. However, Synthea relies on predefined modules and heuristic rules, which may limit its ability to capture complex or atypical clinical scenarios and provide the fine-grained control required to stress-test model reasoning. Recent approaches using LLMs for text-to-FHIR transformation \citep{li_fhir-gpt_2024, frei_infherno_2026} offer improved patient-level control; however, they primarily focus on faithful reconstruction of existing clinical records rather than generating diverse evaluation datasets. 

Taken together, these limitations highlight a key gap: existing approaches do not provide flexible and controllable methods for generating terminology-grounded, FHIR-formatted patient cases that can systematically evaluate model reasoning under diverse and challenging conditions.

To address this gap, we introduce a reusable pipeline for generating synthetic HL7 FHIR R4 patient bundles from unstructured text, with an emphasis on controllability and downstream evaluation. A central component of the pipeline is a terminology-grounded validation and repair step that identifies and corrects hallucinated clinical codes against standard clinical terminologies, while enforcing structural and semantic consistency across generated FHIR resources. This enables interoperable and scalable evaluation of LLM-based clinical systems. 

We further introduce \textbf{MedCase-Structured\footnote{\href{https://huggingface.co/datasets/system-technologies/MedCase-Structured}{Dataset available on Hugging Face}.}}, a structured diagnostic reasoning dataset constructed by applying our pipeline to MedCaseReasoning \citep{wu_medcasereasoning_2025}. \textbf{MedCase-Structured} encodes each case as a terminology-validated FHIR R4 patient bundle, preserving the diagnostic complexity of the original narratives while representing it in an interoperable format. The dataset is designed to evaluate structured diagnostic reasoning over FHIR-style clinical context, with the source case reports providing diagnostically rich and challenging presentations.

\section{Related Work}

\textbf{{Clinical data transformation and interoperability.}}
Prior work focuses on transforming heterogeneous clinical data into standardized formats such as FHIR. Traditional approaches rely on rule-based NLP systems \cite{wang_clinical_2018}, often combining multiple tools for entity extraction and normalization. More recent LLM-based methods, including FHIR-GPT \citep{li_fhir-gpt_2024} and Infherno \citep{frei_infherno_2026}, convert clinical text into structured FHIR resources. However, these approaches primarily reconstruct existing clinical records and remain limited in resource coverage, rather than generating diverse or controllable patient data for downstream evaluation. Synthetic generators such as Synthea \citep{walonoski_synthea_2018} provide large-scale FHIR-compatible patient data, but offer limited control over clinical complexity and patient-level variation.

\textbf{{LLMs for structured EHR and FHIR-based reasoning.}}
Benchmarks such as EHRStruct \citep{yang_ehrstruct_2026} and FHIR-AgentBench \citep{lee_fhir-agentbench_2025} evaluate LLMs on structured EHR and FHIR-based tasks, showing that models struggle with knowledge-driven reasoning, retrieval over complex patient records, and sensitivity to input formats and evaluation settings. However, these benchmarks operate on fixed datasets and cannot generate new patient scenarios, vary clinical complexity, or systematically probe model behavior under controlled conditions.

To our knowledge, no prior work provides controllable, text-driven generation of clinically realistic FHIR records designed specifically for evaluating diagnostic reasoning. Our work addresses this gap by enabling on-demand generation of structured patient data from unstructured inputs for evaluation in clinically realistic settings.

\section{Method}

Generating FHIR from free text using LLMs can produce hallucinated or invalid clinical codes, structurally inconsistent resources, and diagnostic leakage that biases evaluation. We address these issues with a multi-stage synthetic patient generator that converts unstructured English free-text into structurally valid, terminology-grounded HL7 FHIR R4 patient bundles. The generator targets resources needed to represent diagnostic case context rather than attempting full longitudinal EHR reconstruction.

Unlike agent-based approaches in which the model dynamically decides when to invoke tools \citep{frei_infherno_2026}, our pipeline calls LLMs at fixed stages: clinical information extraction, candidate reranking during terminology grounding, FHIR synthesis, and semantic leak detection. These stages are supported by deterministic vocabulary retrieval, structural and clinical-consistency validation, and rule-based post-processing. The extraction model identifies concepts from free-text without emitting clinical codes. Code assignment is deferred to a grounding stage that embeds, retrieves, and reranks candidates against target terminologies. This separation enables extraction validation, terminology grounding, and completeness checks on a flat intermediate representation before FHIR synthesis. After FHIR synthesis, validation errors are fed back through a repair loop, while post-processing handles completeness and normalization.

We use different LLMs for each stage: GPT-5.4-mini \citep{noauthor_gpt_5_4_mini} for extraction, Gemini 3.5 Flash \citep{noauthor_gemini_3_5_flash} for candidate reranking, and Anthropic's Claude (claude-sonnet-4-5) \citep{noauthor_claude_sonnet_4_5} for FHIR synthesis and semantic leak detection. All stages use greedy decoding (temperature 0). Extraction and reranking use JSON-schema-constrained outputs, while synthesis and leak detection use token caps of 64K and 4K, respectively. Reasoning parameters use provider defaults, and extraction and reranking responses are cached to reduce residual nondeterminism.

\subsection{Extraction}
The first LLM stage parses each free-text case into a typed intermediate representation of the patient. It captures demographics and the clinical findings, including symptoms, findings, vitals, labs, medications, procedures, and history, with a verbatim source quote retained for every extracted item. Resource-specific attributes mentioned in prose, such as analyte for laboratory results or dose for medications, are also extracted. Pertinent negatives are flagged so they are not later asserted as positive findings, and the primary diagnosis is captured separately to support optional diagnosis hiding. 

\subsection{Terminology Grounding}
Each extracted concept is routed by type to one or more target terminologies seen in \cref{tab:terminology_mapping}: SNOMED CT \citep{snomed}, LOINC \citep{mcdonald_loinc_2003}, RxNorm \citep{nelson_normalized_2011}, and CVX \citep{cvx}. Codes are drawn from an internally curated terminology store that aggregates OMOP \citep{voss_feasibility_2015} and other interoperable standards.

We first embed each concept description with SapBERT embeddings \citep{liu_self-alignment_2021} and retrieve nearest-neighbor candidates from the routed terminologies using a FAISS index \cite{johnson_billion-scale_2017} of preferred-term embeddings. An LLM reranker then selects the best candidate or rejects all candidates, using the verbatim source quote as context. For example, a quantitative laboratory value such as "serum potassium 3.2 mmol/L" causes a like-named code to be rejected when it has the wrong concept type, as shown in \cref{fig:pipeline}B. Deterministic checks enforce consistency between the selected code and the target FHIR resource, following FHIR R4 coding guidance \citep{noauthor_FHIR_terminology-module}.  A code-system preference selects the expected terminology for each resource, and a semantic-axis constraint rejects codes from incompatible SNOMED CT \citep{snomed} hierarchy, as shown in \cref{tab:terminology_mapping}. Concepts with no acceptable candidate remain uncoded, with the verbatim source quote carried as free-text. A final validation pass checks emitted codes against the curated terminology store and reduces unverifiable codes to text only.

\begin{table}[h]
\centering
\caption{Code-system preference and accepted SNOMED CT axis per FHIR resource. Target terminologies include SNOMED CT \citep{snomed}, LOINC \citep{mcdonald_loinc_2003}, RxNorm \citep{nelson_normalized_2011}, and CVX \citep{cvx}. Codes from other hierarchies are rejected and re-picked; ``$>$'' denotes
fallback and ``---'' an unconstrained axis.}
\label{tab:terminology_mapping}
\footnotesize
\setlength{\tabcolsep}{3.0pt}
\renewcommand{\arraystretch}{1.05}
\begin{tabular}{@{}>{\raggedright\arraybackslash}p{0.22\linewidth}>{\raggedright\arraybackslash}p{0.43\linewidth}p{0.3\linewidth}@{}}
\toprule
\textbf{Vocabulary} & \textbf{FHIR resource} & \textbf{Accepted SNOMED axis}\\
\midrule
SNOMED CT            & \texttt{Condition}          & disorder, finding, situation, morphologic abnormality \\
SNOMED CT            & \texttt{Observation} (symptom/finding)& finding, disorder, morphologic abnormality \\
LOINC $>$ SNOMED CT  & \texttt{Observation} (lab/test)& finding, disorder, morphologic abnormality, observable entity\\
SNOMED CT $>$ LOINC  & \texttt{Observation} (social) & social context\\
SNOMED CT            & \texttt{Procedure}          & procedure \\
RxNorm               & \texttt{MedicationRequest}  & --- \\
CVX                  & \texttt{Immunization}       & --- \\
 SNOMED CT& \texttt{AllergyIntolerance}&---\\
SNOMED CT& \texttt{FamilyMemberHistory}& ---\\
LOINC& \texttt{DiagnosticReport}& ---\\
\bottomrule
\end{tabular}

\end{table}

\subsection{FHIR Synthesis and Validation}
The second LLM stage assembles the grounded intermediate representation into HL7 FHIR R4 resources. The prompt specifies the mapping from each scenario field to its resource type and attributes, as shown in \cref{tab:terminology_mapping}, to maintain structural conformance and clinical consistency. Structural resources for \texttt{Patient} and \texttt{Encounter} are also constructed. Each generated resource is checked by a three-layer validator: FHIR R4 schema conformance, FHIR invariants, and formatting and consistency checks. Validation errors are returned to the synthesis model for up to three repair attempts. A final deterministic post-processing pass enforces missing resources and normalizes units, dates, and status fields.

\subsection{Diagnosis Hiding}
We support configurable suppression of diagnostic conclusions during bundle generation. The assembled bundle is filtered using one of four modes: NONE removes all diagnostic conclusions; HIDDEN removes only the primary diagnosis; EXPLICIT retains only patient-stated conditions; FULL retains all extracted diagnoses. In NONE and HIDDEN modes, code- and substring-based filtering is followed by an LLM semantic scan over all narrative fields to redact residual diagnostic context, including abbreviations, implied conclusions, and synonyms absent from the synonym list.

\section{MedCase-Structured}

In this section, we introduce \textbf{MedCase-Structured}, a synthetic, case-report-derived FHIR dataset for structured diagnostic reasoning. 

\subsection{Dataset}

\begin{table}[t]
\centering
\caption{MedCaseReasoning \citep{wu_medcasereasoning_2025} conversion outcomes across dataset splits. Cases are partitioned into successfully generated MedCase-Structured examples, bundles generated but missing extracted element, and exclusion classes: imaging-dependent cases, and non-human / multi-patient reports.}
\label{tab:dataset_summary}
\small
\setlength{\tabcolsep}{3.0pt}
\renewcommand{\arraystretch}{1.05}
\begin{tabular}{lrrrr@{\hspace{8pt}}r}
\toprule
\textbf{Split} & \textbf{\shortstack{Original\\Total}} &
\textbf{\shortstack{Imaging\\Excluded}} &
\textbf{\shortstack{Subject\\Excluded}}&
\textbf{\shortstack{Incomplete\\Bundles}}&
\textbf{\shortstack{Success\\Cases}}\\
\midrule
Test  & 897      & 769& 17& 1& \textbf{110}\\
Val   & 500      & 428& 7& 2& \textbf{63}\\
Train & 13{,}092 & 11{,}262& 222& 49& \textbf{1{,}559}\\
\bottomrule
\end{tabular}
\end{table}

 The dataset is constructed by applying our reusable text-to-FHIR pipeline to MedCaseReasoning \citep{wu_medcasereasoning_2025}, an open-access dataset of approximately 14,500 diagnostic cases sourced from publicly available case reports and designed to evaluate LLM alignment with clinician-authored diagnostic reasoning. Each case includes a final diagnosis. The original dataset is split into 13,092 training, 500 validation, and 897 test cases.

We exclude cases that are non-human, involve multiple patients, or reference imaging details, which are not currently supported by the generator. The remaining 1,784 cases are processed through the pipeline. This conservative filtering reduces coverage but ensures that released cases satisfy the structural, terminology-grounding, and diagnosis-hiding constraints required for evaluation. 

\cref{tab:dataset_summary} shows the final dataset statistics. Of the pipeline-processed cases, 1,732 are successfully converted into valid FHIR representations (97.1\% success rate).

\subsection{Pipeline Failure Modes}

Among the 1,732 shipped bundles, terminology coverage remains the primary limitation. 18.2\% of distinct extracted clinical concepts (3,775/20,794) were preserved as text but did not receive a normalized code, reflecting code-retrieval limitations concentrated in the categories shown in \cref{tab:coding_gaps}. 

Separately, 52 attempted cases (2.9\% of the 1,784 processed cases) did not produce complete bundles and were excluded from the final \textbf{MedCase-Structured} output. In each, a clinical element extracted from the narrative was dropped during grounding or validation. In 42 cases, the element could not be grounded to a code whose semantic axis matches its FHIR resource type and was suppressed. In the remaining 10, an assigned code failed FHIR or terminology validation.
\begin{table}[t]
\centering
\caption{Terminology coverage in MedCase-Structured. Target terminologies include SNOMED CT \citep{snomed}, LOINC \citep{mcdonald_loinc_2003}, RxNorm \citep{nelson_normalized_2011}, and CVX \citep{cvx}. 
Resource-level coverage is the fraction of code-eligible resource
instances carrying at least one terminology code; concept-level coverage is the fraction of distinct concepts (deduplicated by display text) carrying a code; distinct codes counts unique codes emitted.}
\label{tab:code_coverage}
\small
\setlength{\tabcolsep}{3.0pt}
\renewcommand{\arraystretch}{1.05}
\begin{tabular}{lrrrrr}
\toprule
 & \multicolumn{2}{c}{\textbf{Resource-level}} & \multicolumn{2}{c}{\textbf{Concept-level}} & \\
\cmidrule(lr){2-3}\cmidrule(lr){4-5}
\textbf{Terminology}& $N$ & Coverage & $N$ & Coverage & \textbf{\shortstack{Distinct\\Codes}}\\
\midrule
SNOMED CT & 19{,}517 & 85.0\% & 15{,}573 & 83.5\% & 4{,}619 \\
LOINC     & 15{,}562 & 90.0\% &  3{,}903 & 77.6\% & 1{,}753 \\
RxNorm    &  3{,}681 & 84.7\% &  1{,}280 & 74.5\% &   634 \\
CVX       &     46   & 93.5\% &     38   & 94.7\% &    24 \\
\midrule
\textbf{Overall} & \textbf{38{,}806} & \textbf{87.0\%} & \textbf{20{,}794} & \textbf{81.8\%} & \textbf{7{,}030} \\
\bottomrule
\end{tabular}
\end{table}
\begin{table}[t]
\centering
\caption{Sources of uncoded (text-only) concept gaps. Counts are distinct never-coded concepts; percentages are of unique concepts in each terminology (denominators in Table \ref{tab:code_coverage}). Target terminologies include SNOMED CT \citep{snomed}, LOINC \citep{mcdonald_loinc_2003}, RxNorm \citep{nelson_normalized_2011}, and CVX \citep{cvx}. }
\label{tab:coding_gaps}
\small
\setlength{\tabcolsep}{3.0pt}
\renewcommand{\arraystretch}{1.05}
\begin{tabular}{l r r p{0.45\linewidth}}
\toprule
\textbf{Terminology} & \textbf{\shortstack{Never-\\coded}} &
\textbf{\shortstack{\% of\\unique}} & \textbf{Dominant gap categories}\\
\midrule
SNOMED CT & 2{,}571 & 16.5\% &
Dermatologic/morphologic findings; normal exam findings; axis-mismatched symptoms\\
LOINC & 875 & 22.4\% &
Specialty examinations;
pathology stains; qualitative
serologies\\
RxNorm & 327 & 25.5\% &
Drug-class mentions without an ingredient;
non-RxNorm substances; missing ingredients\\
CVX & 2 & 5.3\% & Negligible\\
\midrule
\textbf{Total} & \textbf{3{,}775} & \textbf{18.2\%} & \\
\bottomrule
\end{tabular}
\end{table}

\subsection{Evaluation}
\begin{table}[t]
\caption{Comparison of LLM diagnostic accuracy on the FHIR-based MedCase-Structured (MCS) dataset and the subset of the corresponding questions in the text-based MedCaseReasoning (MCR) \cite{wu_medcasereasoning_2025}. Evaluated models are GPT-5.4 \citep{noauthor_gpt_5_4}, Gemini-3.1-Pro \citep{noauthor_gemini_3_1_pro}, and Claude-Opus-4.6 \citep{noauthor_claude_opus_4_6}. $\Delta$ is the accuracy difference (MCS $-$ MCR).}
\label{tab:results}
\centering
\footnotesize
\begin{tabular*}{\columnwidth}{@{\extracolsep{\fill}}lccc@{}}
\toprule
Model & MCR & \textbf{MCS} & $\Delta$ \\
\midrule
GPT-5.4 & & & \\
\quad \textit{w/ zero-shot} & 69.39 & 57.37 & \textit{--12.02} \\
\quad \textit{w/ 1-shot}    & 67.27 & 55.45 & \textit{--11.82} \\
\quad \textit{w/ 5-shot}    & 70.91 & 58.18 & \textit{--12.73} \\
\addlinespace
Gemini-3.1-Pro & & & \\
\quad \textit{w/ zero-shot} & 53.03 & 42.42 & \textit{--10.61} \\
\quad \textit{w/ 1-shot}    & 48.18 & 45.45 & \textit{--2.73} \\
\quad \textit{w/ 5-shot}    & 46.36 & 40.91 & \textit{--5.45} \\
\addlinespace
Claude-Opus-4.6 & & & \\
\quad \textit{w/ zero-shot} & 61.82 & 54.55 & \textit{--7.27} \\
\quad \textit{w/ 1-shot}    & 61.82 & 56.82 & \textit{--5.00} \\
\quad \textit{w/ 5-shot}    & 61.37 & 56.30 & \textit{--5.07} \\
\bottomrule
\end{tabular*}
\end{table}

We evaluate the diagnostic accuracy of popular LLM models on \textbf{MedCase-Structured} and compare it to that on the corresponding questions in text format from the original MedCaseReasoning dataset. We average the results over 3 runs and report the mean accuracy. The detailed setup of the experiment is shown in \cref{app:experiment}.

\cref{tab:results} shows the results of our evaluation. LLMs perform consistently worse in diagnostic reasoning when dealing with structured FHIR inputs compared to plain text patient descriptions. We also see insignificant improvement from in-context learning, showing that the ability to understand and reason with structured FHIR data might not be easily learned through simple few shot examples. The experiment results indicate that diagnostic reasoning on structured EHR data is a far more challenging task than simple text-based reasoning.

\section{Conclusion}

We introduce a reusable pipeline to convert clinical case narratives into terminology-grounded HL7 FHIR R4 bundles and release \textbf{MedCase-Structured}, a synthetic FHIR dataset built for evaluating CDSS in structured, EHR-congruent diagnostic reasoning settings.

Our results show that LLMs achieve consistently lower diagnostic accuracy when operating over structured FHIR inputs compared to plain text descriptions. This suggests that structured FHIR inputs may introduce additional challenges for diagnostic reasoning. These findings highlight the importance of evaluating CDSS on deployment-aligned data formats, as performance on simplified or unrelated inputs may not reflect behavior in clinical environments.

Our pipeline has several limitations. It currently supports a limited subset of FHIR resources and represents temporal information through repeated, date-aware resources rather than full longitudinal patient trajectories. Because imaging evidence is not yet supported, the released dataset emphasizes cases whose diagnostic context can be represented in text-grounded FHIR resources. Terminology grounding also remains challenging when clinical descriptions are unsupported, ambiguous, or too specific to map cleanly to a single standardized concept. Future work should broaden terminology retrieval, add axis-aware grounding, extend resource and longitudinal modeling, and incorporate context-aware validation and imaging support to improve robustness.

\section*{Impact Statement}

This work aims to improve evaluation of CDSS in EHR-native settings by generating structured, clinically realistic synthetic patient data for controlled and interoperable benchmarking.

Synthetic data may not fully capture real-world complexity, and errors in generation or terminology grounding may propagate into downstream evaluations. These datasets should therefore complement, not replace, real-world clinical validation.

\section*{Acknowledgment}

We thank David Kang for helpful discussions and feedback on early drafts of this paper. We are grateful to the anonymous reviewers for their constructive comments. This work was supported by System Inc.

\bibliography{references}
\bibliographystyle{icml2026}

\newpage
\appendix
\onecolumn
\section{\textbf{MedCase-Structured} Sample}
\label{app:example}

Listing~\ref{lst:fhir-example} shows the truncated version of a representative structured patient bundle from MedCase-Structured. The full dataset is available at 
\url{https://huggingface.co/datasets/system-technologies/MedCase-Structured}.

\begin{lstlisting}[caption={Example FHIR R4 Patient Bundle from MedCase-Structured.},label={lst:fhir-example}]
{
  "resourceType": "Bundle",
  "type": "collection",
  "entry": [
    {
      "fullUrl": "urn:uuid:d0597131-f20e-47b3-96c4-0c37d5998555",
      "resource": {
        "resourceType": "Patient",
        "id": "d0597131-f20e-47b3-96c4-0c37d5998555",
        "name": [
          {
            "use": "official",
            "given": [
              "Synthetic"
            ],
            "family": "Patient"
          }
        ],
        "gender": "male",
        "birthDate": "1974-01-15"
      }
    },
    {
      "fullUrl": "urn:uuid:17116138-6e0a-4ed3-8235-c9c93ad129d9",
      "resource": {
        "resourceType": "Encounter",
        "id": "17116138-6e0a-4ed3-8235-c9c93ad129d9",
        "status": "finished",
        "class": {
          "system": "http://terminology.hl7.org/CodeSystem/v3-ActCode",
          "code": "AMB",
          "display": "ambulatory"
        },
        "type": [
          {
            "coding": [
              {
                "system": "http://snomed.info/sct",
                "code": "185347001",
                "display": "Encounter for problem"
              }
            ],
            "text": "Encounter for problem"
          }
        ],
        "subject": {
          "reference": "Patient/d0597131-f20e-47b3-96c4-0c37d5998555"
        },
        "period": {
          "start": "2026-06-25T09:00:00-05:00",
          "end": "2026-06-25T10:00:00-05:00"
        },
        "reasonCode": [
          {
            "coding": [
              {
                "system": "http://snomed.info/sct",
                "code": "70819003",
                "display": "Erythema"
              }
            ],
            "text": "Severe redness around right elbow"
          }
        ]
      }
    },
    {
      "fullUrl": "urn:uuid:bc787ee0-4ff6-4260-a49c-ff48d9154d9b",
      "resource": {
        "resourceType": "Condition",
        "id": "bc787ee0-4ff6-4260-a49c-ff48d9154d9b",
        "clinicalStatus": {
          "coding": [
            {
              "system": "http://terminology.hl7.org/CodeSystem/condition-clinical",
              "code": "active",
              "display": "Active"
            }
          ],
          "text": "Active"
        },
        "verificationStatus": {
          "coding": [
            {
              "system": "http://terminology.hl7.org/CodeSystem/condition-ver-status",
              "code": "confirmed",
              "display": "Confirmed"
            }
          ]
        },
        "category": [
          {
            "coding": [
              {
                "system": "http://terminology.hl7.org/CodeSystem/condition-category",
                "code": "problem-list-item",
                "display": "Problem List Item"
              }
            ],
            "text": "Problem List Item"
          }
        ],
        "code": {
          "coding": [
            {
              "system": "http://snomed.info/sct",
              "code": "70819003",
              "display": "Erythema"
            }
          ],
          "text": "Severe redness around right elbow"
        },
        "bodySite": [
          {
            "coding": [
              {
                "system": "http://snomed.info/sct",
                "code": "368148009",
                "display": "Right elbow region"
              }
            ],
            "text": "right elbow"
          }
        ],
        "subject": {
          "reference": "Patient/d0597131-f20e-47b3-96c4-0c37d5998555"
        },
        "onsetDateTime": "2026-06-18",
        "recordedDate": "2026-06-25",
        "severity": {
          "text": "severe"
        }
      }
    },
    {
      "fullUrl": "urn:uuid:3235971f-df4a-483a-8134-6e34f3bd8103",
      "resource": {
        "resourceType": "Condition",
        "id": "3235971f-df4a-483a-8134-6e34f3bd8103",
        "clinicalStatus": {
          "coding": [
            {
              "system": "http://terminology.hl7.org/CodeSystem/condition-clinical",
              "code": "active",
              "display": "Active"
            }
          ],
          "text": "Active"
        },
        "verificationStatus": {
          "coding": [
            {
              "system": "http://terminology.hl7.org/CodeSystem/condition-ver-status",
              "code": "confirmed",
              "display": "Confirmed"
            }
          ]
        },
        "category": [
          {
            "coding": [
              {
                "system": "http://terminology.hl7.org/CodeSystem/condition-category",
                "code": "problem-list-item",
                "display": "Problem List Item"
              }
            ],
            "text": "Problem List Item"
          }
        ],
        "code": {
          "coding": [
            {
              "system": "http://snomed.info/sct",
              "code": "90673000",
              "display": "Burning sensation"
            }
          ],
          "text": "Intense burning sensation of right elbow"
        },
        "bodySite": [
          {
            "coding": [
              {
                "system": "http://snomed.info/sct",
                "code": "368148009",
                "display": "Right elbow region"
              }
            ],
            "text": "right elbow"
          }
        ],
        "subject": {
          "reference": "Patient/d0597131-f20e-47b3-96c4-0c37d5998555"
        },
        "onsetDateTime": "2026-06-18",
        "recordedDate": "2026-06-25",
        "severity": {
          "text": "intense"
        }
      }
    },
    {
      "fullUrl": "urn:uuid:3f5011c0-ee84-4a74-8541-ddb995bb8ddc",
      "resource": {
        "resourceType": "Condition",
        "id": "3f5011c0-ee84-4a74-8541-ddb995bb8ddc",
        "clinicalStatus": {
          "coding": [
            {
              "system": "http://terminology.hl7.org/CodeSystem/condition-clinical",
              "code": "active",
              "display": "Active"
            }
          ],
          "text": "Active"
        },
        "verificationStatus": {
          "coding": [
            {
              "system": "http://terminology.hl7.org/CodeSystem/condition-ver-status",
              "code": "confirmed",
              "display": "Confirmed"
            }
          ]
        },
        "category": [
          {
            "coding": [
              {
                "system": "http://terminology.hl7.org/CodeSystem/condition-category",
                "code": "problem-list-item",
                "display": "Problem List Item"
              }
            ],
            "text": "Problem List Item"
          }
        ],
        "code": {
          "coding": [
            {
              "system": "http://snomed.info/sct",
              "code": "403599000",
              "display": "Stinging of skin"
            }
          ],
          "text": "Intense stinging sensation of right elbow"
        },
        "bodySite": [
          {
            "coding": [
              {
                "system": "http://snomed.info/sct",
                "code": "368148009",
                "display": "Right elbow region"
              }
            ],
            "text": "right elbow"
          }
        ],
        "subject": {
          "reference": "Patient/d0597131-f20e-47b3-96c4-0c37d5998555"
        },
        "onsetDateTime": "2026-06-18",
        "recordedDate": "2026-06-25",
        "severity": {
          "text": "intense"
        }
      }
    },
// ... additional resources omitted for brevity
}
\end{lstlisting}

\section{Experimental Setup}
\label{app:experiment}

We use commercialized API endpoints provided by OpenAI, Google, and Anthropic to prompt corresponding LLMs for diagnostic reasoning on MedCaseReasoning and MedCase-Structured. We set reasoning parameters to medium, max generation tokens to 800, and temperature to 1.0 across all experiments. For few-shot learning cases, we randomly sample cases from the training split to build the few shot learning prompts for each run.

For evaluation, we use a Claude Sonnet 4.6 model as the LLM judge to compare the "diagnosis" field to the ground truth diagnosis string. We prompt the judge to assess whether the predicted diagnosis is clinically equivalent to the ground truth and output a final binary decision.

\subsection{Diagnostic Reasoning Prompt}

\begin{prompt}[System Prompt]
You are a careful physician solving clinical diagnostic reasoning cases. Use only the provided case information. Return valid JSON only.
\end{prompt}

\vspace{0.5em}

\begin{prompt}[User Prompt]
You will receive a \{\textit{FHIR Bundle JSON for a clinical case} OR \textit{plain text clinical case description}\}. Determine the most likely final diagnosis.

Return exactly this JSON schema:
{{
  "diagnosis": "single most likely diagnosis",
  "reasoning": "brief explanation using the case evidence"
}}

\texttt{\{fhir\_bundle\}}

Now solve the target case.

Target case:
\texttt{\{case\_input\}}
\end{prompt}

\subsection{Judge Prompt}

\begin{prompt}[System Prompt]
You judge whether a predicted diagnosis is clinically equivalent to the ground truth. Accept synonyms, spelling variants, and equivalent specificity. Return valid JSON only.
\end{prompt}

\vspace{0.5em}

\begin{prompt}[User Prompt]
Ground truth diagnosis:
\texttt{\{ground\_truth\}}

Predicted diagnosis:
\texttt{\{prediction\}}
\end{prompt}


\end{document}